\pdfoutput=1

\documentclass[11pt]{article}

\usepackage[]{acl}

\usepackage{times}
\usepackage{latexsym}

\usepackage[T1]{fontenc}

\usepackage[utf8]{inputenc}
\usepackage{natbib}
\usepackage{microtype}
\usepackage{booktabs}
\usepackage{subcaption}
\usepackage{url}
\usepackage{comment}
\usepackage{todonotes}
\usepackage{color,soul}
\usepackage{graphicx}
\usepackage[normalem]{ulem}
\usepackage{algorithm}
\usepackage{algorithmic}
\usepackage{amsmath,amssymb}
\usepackage{longtable}
\usepackage{lscape}
\usepackage{enumitem}
\usepackage[normalem]{ulem}
\useunder{\uline}{\ul}{}

\title{TeamX@DravidianLangTech-ACL2022: A Comparative Analysis for Troll-Based Meme Classification}

\author{Rabindra Nath Nandi$^1$, Firoj Alam$^2$, Preslav Nakov$^2$\\
  $^1$BJIT Limited, Dhaka, Bangladesh \\
  $^2$Qatar Computing Research Institute, HBKU, Doha, Qatar \\
  \texttt{rabindra.nath@bjitgroup.com},
  \texttt{\{falam,pnakov\}@hbku.edu.qa}
  \\}

\begin{document}
\maketitle
\begin{abstract}
The spread of fake news, propaganda, misinformation, disinformation, and harmful content online raised concerns among social media platforms, government agencies, policymakers, and society as a whole. This is because such harmful or abusive content leads to several consequences to people such as physical, emotional, relational, and financial. Among different harmful content \textit{trolling-based} online content is one of them, where the idea is to post a message that is provocative, offensive, or menacing with an intent to mislead the audience. The content can be textual, visual, a combination of both, or a meme. In this study, we provide a comparative analysis of troll-based memes classification using the textual, visual, and multimodal content. We report several interesting findings in terms of code-mixed text, multimodal setting, and combining an additional dataset, which shows improvements over the majority baseline. 
\end{abstract}
\section{Introduction}
\label{sec:intro}
Social media have become one of the main communication channels for the propagation of information through textual, visual, or audio-visual content. While the content shared on social media creates a positive impact, however, there are also content that spread harm and hostility \citep{brooke-2019-condescending}, 
including 
abusive language \citep{mubarak2017abusive}, propaganda \cite{da2020survey,EMNLP19DaSanMartino}
cyberbullying \citep{van-hee-etal-2015-detection}, 
cyber-aggression \citep{kumar2018benchmarking}, and other kinds of harmful content \citep{pramanick-etal-2021-momenta-multimodal}.
The propagation of such content most often is done by an automated tool, troll, or coordinated groups, which target specific users, communities (e.g., minority groups), individuals, and companies. To detect such content there has been effort to develop automatic tools (see most recent surveys on 
disinformation \citep{alam2021survey}, rumours \citep{bondielli2019survey}, propaganda \citep{da2020survey}, and multimodal memes  \citep{afridi2021multimodal}, hate speech \citep{fortuna2018survey}, 
cyberbullying \citep{7920246}, and offensive content \citep{husain2021survey}. In addition, shared tasks has also been organized in the past years addressing factuality, fake news and harmful content~\cite{clef-checkthat:2021:LNCS,kiela2020hateful}. 

Among other social media content, recently, the uses of \textit{internet memes} became popular and they are often shared for the purpose of humor or fun with no bad intentions. However, memes are also created and shared with harmful intentions. This include attack on people based on the characteristics such as ethnicity, race, sex, gender identity, disability, disease, nationality, and immigration status \cite{kiela2020hateful}.
There has been research effort to develop computational method to detect such memes, such as detecting hateful memes \cite{kiela2020hateful}, propaganda \cite{dimitrov2021detecting,SemEval2021-6-Dimitrov}, offensive \cite{suryawanshi-etal-2020-multimodal}, sexist meme \cite{fersini2019detecting} and troll based meme \cite{dravidiantrollmeme-eacl}. 

In this study, we focus on troll-based meme classification based on the dataset released in the shared task discussed in \cite{trollmeme-eacl}. While meme contains both textual and visual elements, hence, we investigate textual, visual content and their combination
using different pretraind transformer models.
In addition, we explored combining an external dataset, and use of code-mixed text (i.e., Tamil and English) extracted using OCR. Note that the text provided with the dataset is transcribed in Latin. While prior work focuses on the text provided with the dataset, here, we also follow a different strategy, directly using the text from the OCR without any cleaning. 

Our contributions include:
\begin{itemize} 
\item 
we investigate classical algorithm (e.g., SVM), pretraind transformer and deep CNN models for both text and images, respectively;
\item we combine an additional dataset and use code-mixed text, extracted using OCR and compare the performance;  
\item we also experiment with different pretrained multimodal models. 
\end{itemize}
    
\section{Related Work}
\label{sec:related_work}
Prior work on detecting harmful aspects of memes include categorizing hateful memes \cite{kiela2020hateful}, antisemitism \cite{chandra2021subverting} and propaganda detection techniques in memes \cite{dimitrov2021detecting}, harmful memes and their target \cite{pramanick-etal-2021-momenta-multimodal}, identifying protected category such as race, sex  that has been attacked~\cite{zia-etal-2021-racist}, and identifying offensive content \cite{suryawanshi-etal-2020-multimodal}. Among the studies most notable effort that streamlined the research work include shared tasks such as ``Hateful Memes Challenge'' \cite{kiela2020hateful}, detection of persuasion techniques \cite{SemEval2021-6-Dimitrov} and troll meme classification \cite{dravidiantrollmeme-eacl}. 

The work by \citet{chandra2021subverting} investigates antisemitism along with its types by addressing the tasks as binary and multi-class classification using  pretrained transformers and CNN as modality-specific encoders along with various multimodal fusion strategies. \citet{dimitrov2021detecting} developed a dataset with 22 propaganda techniques and investigates the different state-of-the-art pretrained models and demonstrate that joint vision-language models perform best. \citet{pramanick-etal-2021-momenta-multimodal} address two tasks such as detecting harmful memes and identifying the social entities they target and propose a multimodal model, which utilizes local and global information. \citet{zia-etal-2021-racist} goes one step further than a binary classification of hateful memes -- more fine-grained categorization based on protected category (i.e., race, disability, religion, nationality, sex) and their attack types (i.e., contempt, mocking, inferiority, slurs, exclusion, dehumanizing, inciting violence) using the dataset released in the WOAH 2020 Shared Task.\footnote{\url{github.com/facebookresearch/fine_grained_hateful_memes}} 
\citet{fersini2019detecting} studied sexist meme detection and investigate textual cues with a late-fusion strategy, which suggest that fusion approach performs better. The same authors also developed a dataset of 800 misogynistic memes covering different manifestations of hatred against women (e.g., body shaming, stereotyping, objectification and violence), which are collected from different social media~\citep{france_mys2021}.

In the ``Hateful Memes Challenge'', the participants addressed the hateful meme classification task by fine-tuning the state-of-art multi-modal transformer models \citep{kiela2021hateful} and best system in the competition used different unimodal and multimodal pre-training models such as VisualBERT \citep{li2019visualbert} VL-BERT \citep{su2019vl}, UNITER \citep{chen2019uniter}, VILLA \citep{gan2020large} and 
ensembles \citep{kiela2021hateful}. 
The SemEval-2021 propaganda detection shared task \citep{SemEval2021-6-Dimitrov} was organized with a focus on fine-grained propaganda techniques in text and the entire meme, and from the participants' systems, they conclude that multimodal cues are important for automated propaganda detection. In the troll meme classification shared task \cite{dravidiantrollmeme-eacl}, the best system used ResNet152, BERT with multimodal attention, and the majority of the system used pretrained transformer models for text, CNN models for images, and early fusion approaches.

\section{Experiments}
\label{ssec:classification}
\subsection{Data}
\label{ssec:data}
We use the dataset provided in the troll-based Tamil meme classification shared task discussed in \cite{suryawanshi-etal-2020-dataset,trollmeme-eacl}. The dataset is comprised of meme and transcribed text in Latin, which are annotated and transcribed by native Tamil speakers. There is a total of 2,300, 667 memes for training and testing, respectively. For our experiments, we split the training set into training and development set with 80 and 20\%, respectively. The development set is used for fine-tuning the models.

In Table \ref{tab:data_dist}, we report the distribution of the dataset that we used for the experiments. 
\begin{table}[tbh!]
\centering
\begin{tabular}{lrrr}
\toprule
\textbf{Class } & \textbf{Train} & \textbf{Dev} & \textbf{Test} \\ \midrule
Troll & 1013  & 269 & 395 \\
Non-Troll & 827 & 191 & 272 \\
\bottomrule
\end{tabular}
\caption{Distribution of the Troll-based Tamil Meme dataset. We split original training set into training and development set.}
\label{tab:data_dist} 
\end{table}

\subsection{Settings}
\label{ssec:models}
For the classification, we run different unimodal 
experiments: {\em (i)} only text, {\em (ii)} only meme, and {\em (iii)} text and meme together. 
For each setting, we also run several baseline experiments. One such baseline is the \textit{majority class} baseline, which predicts the label based on the most frequent label in the training set. 
This has been most commonly used in shared tasks \cite{clef-checkthat:2021:LNCS}.
Furthermore, we run a few advanced experiments using an additional dataset and code-mixed text from OCR. To measure the performance of each model we used a weighted F$_1$ score to maintain shared task guideline.

\subsubsection{Text Modality}
For the baseline using text modality, we used bag-of-$n$-gram vectors weighted with logarithmic term frequencies (tf) multiplied with inverse document frequencies (idf) and train the model using Support Vector Machines (SVM)~\cite{platt1998sequential33}. Note that we extracted unigram, bigram, and tri-gram features. We used grid search to optimize the SVM hyper-parameters.

We then experiment using multilingual BERT (mBERT) model \cite{text_bert}.
We performed ten reruns for each experiment using different random seeds, then we picked the model that performed best on the development set. We used a batch size of 8, a learning rate of 2e-5, maximum sequence length 128, three epochs, and used the `categorical cross-entropy' as the loss function. 

\subsubsection{Image Modality}
Similar to the text modality, for the baseline experiment with image modality, we extract features from a pre-trained model, then train the model using SVM. We extracted features from the penultimate layer of the EfficientNet (b1) model \cite{tan2019efficientnet}, which was trained using ImageNet. For training the model using SVM we used the default parameters setting.

For the later experiments we used the transfer learning approach, fine-tuning the pre-trained deep CNN models (e.g., VGG16), which has been shown success for visual recognition tasks. 
We used the weights of the model pre-trained on ImageNet to initialize our model. We adapt the last layer (i.e., softmax layer) of the network for the binary classification task. 
We trained models using three popular neural network architectures such as VGG16~\cite{simonyan2014very}, ResNet101~\cite{he2016deep} and EfficientNet~\cite{tan2019efficientnet}, which showed state-of-art performance in similar tasks \cite{multimodalbaseline2020,alam2021medic,alam2021social}. For training, we used the Adam optimizer~\cite{kingma2014adam} with an initial learning rate of $10^{-5}$, which is decreased by a factor of 10 when accuracy on the dev set stops improving for 10 epochs. 
From our experiment, we observe that the model converse within 50-60 epochs. As the dataset size is small, fine-tuning the entire network did not yield better results, therefore, we freeze the network and fine-tune the penultimate layer.

\subsubsection{Multimodal: Text and Image}

For the multimodel experiments, we used Vision Transformer (ViT) \cite{Dosovitskiy2021} for image feature encoding and multilingual BERT (mBERT) for the textual representation. 
We used BLOCK fusion \cite{BlockFusion2019} to merge the features from two different modalities. 
BLOCK fusion is a multimodal fusion based on block-superdiagonal tensor decomposition. Previously, \citet{Dosovitskiy2021} showed a better performance using BLOCK fusion over several bilinear fusion techniques for Visual Question Answering (VQA) and Visual Relationship Detection (VRD) tasks. We conduct two major experiments by varying the textual data, {\em(i)} using the provided text, and {\em(ii)} using the code-mixed text.

\subsubsection{Additional Experiments} \mbox{} \\
\textbf{Code-mixed Tamil and English text:} 
The dataset comes with extracted text in a transcribed form in Latin. Given that current multilingual transformer models have not trained using such a latin form of text, therefore, to further understand the problem we extract text from memes using \textit{tesseract} \cite{smith2007overview}.\footnote{\url{https://pypi.org/project/pytesseract/}} 
We then train the same mBERT model using the extracted code-mixed Tamil and English text. Note that the extracted text contains noise that comes from the output of the OCR. 


\textbf{Additional data:} 
While we visually inspected the dataset we realized that textual and visual elements of memes have similarities with memes in hateful meme dataset \cite{kiela2020hateful}. We then mapped the labels from hateful memes dataset to troll labels, {\em(i)} \textit{hateful} to \textit{troll}, and {\em(ii)} \textit{not-hateful} to \textit{non-troll}. We combined all training and test set memes from the hateful memes dataset with the training set of the troll meme dataset. The dev set of the hateful meme dataset is combined with the dev set of the troll dataset. After combining the datasets we experiment with both text and image modalities using the same models.

\section{Results and Discussion}
\label{sec:discussion}

\begin{table}[]
\centering
\setlength{\tabcolsep}{2.0pt}
\scalebox{0.90}{
\begin{tabular}{lrrrr}
\toprule
\multicolumn{1}{c}{\textbf{Exp}} & \multicolumn{1}{c}{\textbf{Acc}} & \multicolumn{1}{c}{\textbf{P}} & \multicolumn{1}{c}{\textbf{R}} & \multicolumn{1}{c}{\textbf{F1}} \\ \midrule
Maj. & 59.2 & 35.1 & 59.2 & 44.1 \\ \midrule
\multicolumn{5}{c}{\textbf{Text modality}} \\ \midrule
Tf-Idf + SVM & 46.2 & 47.9 & 46.2 & \textbf{46.6} \\
mBERT & 52.6 & 51.0 & 52.6 & \textbf{51.4} \\
Add. data + mBERT & 56.5 & 52.8 & 56.5 & \textbf{51.8} \\
Code-mixed text + mBERT & 57.3 & 55.2 & 57.3 & \textbf{55.2} \\ \midrule
\multicolumn{5}{c}{\textbf{Image modality}} \\ \midrule
EffNet feat +SVM & 59.1 & 55.6 & 59.1 & \textbf{50.1} \\
VGG16 & 55.8 & 50.2 & 55.8 & \textbf{49.2} \\
ResNet101 & 60.4 & 58.6 & 60.4 & \textbf{53.8} \\
EffNet (b1) & 61.5 & 61.5 & 61.5 & \textbf{53.6} \\ 
\midrule
\multicolumn{5}{c}{\textbf{Multimodality}} \\ \midrule
ViT + mBERT & 57.9 & 55.1 & 57.9 & \textbf{54.2} \\
ViT + mBERT (code-mixed text) & 55.3 & 57.0 & 55.3 & \textbf{55.7} \\ \midrule
\multicolumn{5}{c}{\textbf{Image modality + Additional data}} \\ \midrule
VGG16 & 54.4 & 52.0 & 54.4 & \textbf{52.3} \\
ResNet101 & 60.7 & 58.9 & 60.7 & \textbf{56.2} \\
EffNet (b1) & 60.7 & 58.9 & 60.7 & \underline{\textbf{56.6}} \\ 
\bottomrule
\end{tabular}
}
\caption{Evaluation results on the test set. The results that improve over the majority class baseline are in \textbf{bold}, and the best one is \underline{underlined}. Maj.: Majority baseline, Add. data: Additional data.
}
\label{tab:results}
\end{table}

In Table \ref{tab:results}, we present the results for different modalities and settings. Overall, all results are better than majority baseline. In the unimodal experiments, the image-only models perform better than the text-only models. Our experiments on text-only models suggest that code-mixed data help in improving the performance using the same multiligual mBERT model compared to the transcribed latin text, which suggest that multilingual model is not able to capture the information in latin Tamil text. The additional data, which is in English only, slightly improved the performance for text-only experiment. 

For the image-only experiments, we obtain a comparative performance with ResNet101 and EfficientNet (EffNet (b1)). 


Our experiments on multimodel experiments provide better results compared to text and image modalities as shown in the Table \ref{tab:results}.
For the two multimodel experiments, we obtained weighted-F1 scores of 54.2 and 55.7 for text and for code-mixed text, respectively, which are better than the best weighted-F1 score from text modality (51.4) and from image modality (53.8).
Note that, the performance of our multimodel experiments is better than \citet{Siddhanth2021}, where a weighted F1 score of 0.47 has been reported for a similar task. That implies BLOCK fusion has advantages over late fusion for troll meme classification tasks. 

With additional data, results improved significantly for all image-only models. It confirms our observation that visual elements in memes share information across different datasets, and possibly different cultures too.

Our future plan is to apply the data augmentation technique for multimodality experiments and also we have a plan to explore the performance of popular multimodel algorithms.

During the shared task participation, we submitted one run, using only image modality, where our system was not showing promising performance. However, our subsequent experiments and analysis show several promising directions in terms of original code-mixed data and additional data from the other task.

\section{Conclusion and Future Work}
\label{sec:conclutions}
We present a comparative analysis of different modalities for the troll-based meme classification task. We show that the multilingual model capture more information for code-mixed text than its Latin counterpart. We present higher performance with multimodality compared to unimodal models. Our experiments also suggest that additional data from the other task helps in capturing visual information. In the future, we plan to further develop multimodal models that can capture information in code-mixed noisy conditions. 

\bibliography{bib/all_bib,bib/bibliography}
\bibliographystyle{acl_natbib}



\end{document}